\documentclass[tinyml]{acmart}

\AtBeginDocument{%
  \providecommand\BibTeX{{%
    \normalfont B\kern-0.5em{\scshape i\kern-0.25em b}\kern-0.8em\TeX}}}

\setcopyright{rightsretained}
\copyrightyear{2023}
\acmYear{2023}

\usepackage{xspace}
\usepackage{caption}
\usepackage{multirow}
\usepackage{booktabs}
\usepackage[export]{adjustbox}
\usepackage{amsmath}
\usepackage{placeins}
\usepackage{algorithm}
\usepackage{subfigure}
\usepackage{algpseudocode}
\usepackage{bbm}
\usepackage{lipsum}
\usepackage[T1]{fontenc}        
\usepackage[utf8]{inputenc}
\newcommand\Tau{\mathcal{T}}
\usepackage[detect-weight=true, detect-family=true]{siunitx}

\newcommand{\round}[1]{\num[round-mode=places,round-precision=2]{#1}}
\newcommand{\acc}[2]{\round{#1}\ifthenelse{\equal{#2}{}}{}{\tiny ${\scriptstyle \pm}$\round{#2}}}

\begin{document}
\newcommand{\ouralg}{MetaLDC\xspace}
\newcommand{\ouralgfull}{MetaLDC-full\xspace}
\newcommand{\ouralgnft}{MetaLDC-NFT\xspace}

\title{\ouralg: Meta Learning of 
Low-Dimensional Computing Classifiers 
for Fast On-Device Adaption}

\author{Yejia Liu}
\affiliation{%
  \institution{UC Riverside}
  }
\email{yliu807@ucr.edu}

\author{Shijin Duan}
\affiliation{%
  \institution{Northeastern University}
  }
\email{duan.s@northeastern.edu}

\author{Xiaolin Xu}
\affiliation{%
  \institution{Northeastern University}
  }
\email{x.xu@northeastern.edu}

\author{Shaolei Ren}
\affiliation{%
  \institution{UC Riverside}
  }
\email{sren@ece.ucr.edu}

\renewcommand{\shortauthors}{Liu, et al.}

\begin{abstract}
Fast model updates for unseen tasks on intelligent edge devices are crucial
but also challenging due to the limited computational power. 
In this paper, we propose \ouralg, which meta-trains  brain-inspired ultra-efficient
low-dimensional computing classifiers to enable fast adaptation on tiny
devices with minimal computational costs.
Concretely, during the meta-training stage, \ouralg meta trains a representation offline by explicitly taking into
account that the final (binary) class layer will be fine-tuned for fast
adaptation for unseen tasks on tiny devices;
during the meta-testing stage, \ouralg uses closed-form gradients of
the loss function to enable fast adaptation of the class layer.
Unlike traditional neural networks, \ouralg is designed
based on the emerging LDC framework to enable ultra-efficient on-device inference. Our experiments have demonstrated that compared to SOTA baselines,
\ouralg achieves higher accuracy, robustness against random bit errors, as well as cost-efficient hardware computation.
\end{abstract}


\maketitle

\section{Introduction} 
Deep neural networks (DNNs)
have become the backbone of  intelligent applications in a wide range of domains from computer vision to natural language processing~\cite{nlp,dnn,LIU201711}.
Meanwhile, compared to cloud-based inference,
on-device inference has numerous
advantages, including better privacy preservation and anytime inference
without relying on network connections.
Nonetheless, despite the recent progress \cite{efficient, edgesurvey},
directly running DNN inference and adapting the model to  unseen tasks 
on the edge are still challenging due to
the conflict between high computational demand of DNNs and the
low resource availability of edge devices,  especially tiny
devices such as microcontrollers and Internet-of-Things (IoT)
devices.

In response to the excessive resource demand of DNNs, hyperdimensional computing (HDC) has emerged as an alternative towards efficient on-device inference
~\cite{adaptHDC}. 
The key idea of HDC is to encode data into  (binary) hypervectors each with dimensions of thousands or even more, and then perform cosine/Hamming distance similarity 
for inference using bit-wise binary operations in parallel.
Owning to its hardware friendliness and efficiency, 
HDC classifiers have been adopted to an increasingly broader range of 
inference tasks for resource-constrained devices~\cite{duan2022lehdc, Datta2019APH}. 

Nonetheless, there are still
fundamental limitations that prohibit the applicability of HDC
for tiny devices with extremely limited
resources.
First, the orders of megabyte of memory required by  HDC to support its hyperdimensional data representation can  be too costly for tiny devices  ~\cite{Imani2020QuantHDAQ,ldc}.
Second, compared to today's DNNs, the HDC training process is
extremely rudimentary
(e.g., simply taking the average of inputs, plus some semi-blind heuristic
adjustments, without loss functions), resulting in low inference accuracy.
Last but not least, each HDC training process can only fit into
one data distribution, which means that HDC training does not scale
to a large number of tiny devices, each having potentially different distributions.

While the recent brain-inspired low-dimensional computing (LDC) classifiers outperform
HDC by utilizing ultra-low dimensional vectors and principled training based on an equivalent neural network
to improve the inference efficiency and accuracy~\cite{ldc},
it cannot support fast model adaptation to unseen tasks on tiny devices.
More concretely, LDC  trains an individual model 
for each device, and hence the total training
cost can be labor-intensive when there are many tiny devices deployed
in heterogeneous environments each with
different data distributions\cite{lin2020collaborative}.
Additionally, fast adaption to unseen tasks with only a handful of data points 
presents substantial challenges for tiny devices, 
due to their constrained computing resources that prohibit traditional model updates based on gradients and backpropagation. 
While some studies have proposed to train a collaborative (DNN) model in a distributed manner to facilitate knowledge transfer between edge devices~\cite{lin2020collaborative},
the communication latency among edge nodes and complicated local neural network computation make this approach still too expensive for a tiny device.

\begin{figure*}[!t]
\centering
\includegraphics[width=\textwidth]{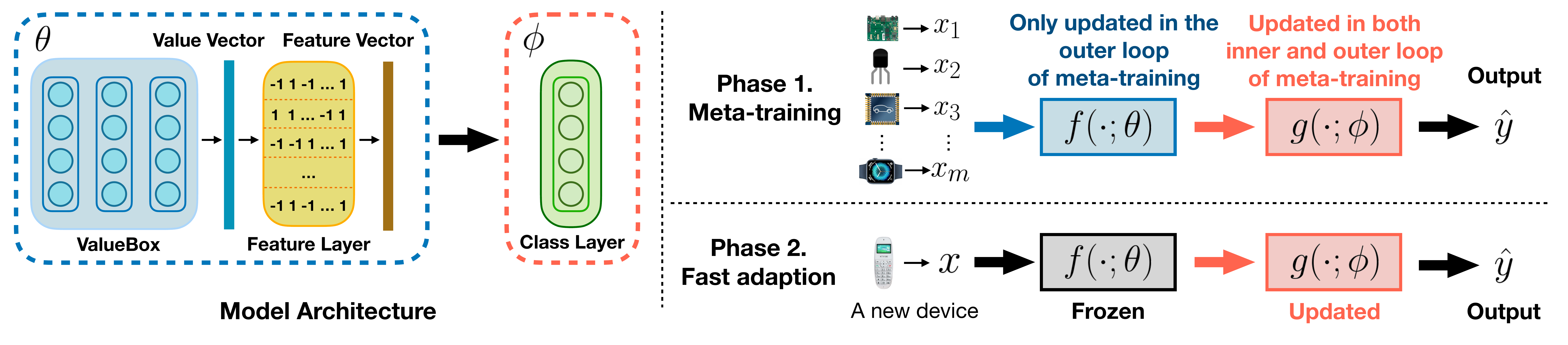}
\caption{Overview of \ouralg. Our \ouralg is based on the ultra-lightweight LDC network and composed of two stages: meta-training and fast adaption. The meta parameter $\theta$ keeps learned knowledge from heterogeneous data distributions, while the task-specific parameter $\phi$ is used to quickly adapt to unseen tasks from a new tiny device.
 }
\label{fig:lol}
\end{figure*}


In the presence of heterogeneous tiny devices each
with a different data distribution,
we propose \ouralg, a new LDC-based meta learning approach
that achieves fast model adaptation to an unseen task
and ultra-efficient inference on tiny devices.
Specifically, \ouralg meta trains an LDC-equivalent neural
network, in which the first few layers are non-binary values specifically for
LDC data encoding and the last layer has binary weights (denoted by
$\phi$) for classification.
Crucially, given the difficulty in calculating
gradients and performing backpropagation over the entire
LDC-equivalent neural network on resource-constrained tiny devices,
we only treat the last layer $\phi$ as task-specific:
the LDC data encoding part is explicitly learned
 to fit heterogeneous distributions, while only the classification
 layer $\phi$ is adapted using a simple closed-form gradient expression to each tiny device for fast and inexpensive
 model update with minimal on-device computational costs.
Thus, unlike the standard meta learning technique (e.g., MAML \cite{finn2017modelagnostic}), \ouralg meta trains the first few layers in
the LDC-equivalent neural network by explicitly considering that these layers
will be fixed without updates on tiny devices.

Our experimental results have empirically shown that \ouralg can significantly outperform state-of-the-art (SOTA) baselines including SOTA HDC~\cite{imani} and pretrained LDC model on both vision and non-vision datasets in terms of accuracy, robustness against the random bit errors on hardware, as well as cost efficiency including energy consumption, latency and model size.

\section{Background}\label{sec:background}
\textbf{Hyper-dimensional computing (HDC):} The HDC is a brain-inspired, cognitive computing architecture built on a unique data type, referred to as the hypervectors. The dimensionality of the hypervectors can be from a thousand to tens of thousands. 
By manipulating these hypervectors, the HDC aims to perform cognitive tasks via hardware-efficient operations like element-wise additions and dot products. 

In the general supervised classification based on HDC, the features of the input sample would be encoded into hypervectors $\mathcal{F}_i$s, together with their corresponding value hypervectors $\mathcal{V}_{f_i}$s,
pre-stored in the item memory. By binding $\mathcal{F}$ and $\mathcal{V}$ via:
\begin{equation}
sgn(\sum_i \mathcal F_i \times \mathcal V_{f_i}),
\nonumber
\end{equation}
we can obtain the encoded input sample $\mathcal{H}$. 
In training, all $\mathcal{H}$s belonging to the same class would be summed and averaged to obtain the class hypervectors, stored in the associative memory. In the inference stage, the testing data would be transformed into the query hypervectors using the same encoder. Then a similarity checker like the Hamming distance would be applied in the associative memory between each trained class hypervectors and the query hypervector. The class label with the closest distance would be returned. Due to the simplicity of bitwise operations, the HDC has achieved success on platforms like FPGA and ASIC~\cite{Datta2019APH}. 

However, the large model size resulted from the ultra-high dimensions of the data representation in the HDC compromise its wide adoption on tiny devices, which are usually under severe resource consumption constraints. On the other hand, although numerous endeavors have been put into improving the accuracy of the HDC classifier~\cite{imani}, there is still a large accuracy gap between the HDC models and a simple modern neural network model like the Multi-Layer Perceptron (MLP). 

\textbf{Low-Dimensional Computing (LDC):}
To overcome the fundamental limitations
of low accuracy and inference efficiency in HDC,
the low-dimensional computing (LDC) classifiers are proposed as a brain-inspired substitute of HDC classifiers 
with higher accuracy and order-of-magnitude better on-device inference efficiency, especially for tiny devices with intelligent needs. Unlike HDC, LDC classifiers 
offer a systematic training procedure, where the value vectors $\mathcal{V}$s and feature vectors $\mathcal{F}$s are explicitly optimized rather than being randomly generated. On the other hand, the required order of magnitude of the involved vectors dimension size in the LDC is only a few to tens to achieve a higher accuracy compared to the state-of-the-art HDC, e.g. 87.38\% w/ $D = 8,000$ vs. 91.22\% w/ $D=4/64$  on the MNIST dataset~\cite{ldc}.

The blue and orange boxes with dashed lines in Figure~\ref{fig:lol} provides an overview of the LDC model architecture. The ValueBox is an encoding network. It maps the feature values $\mathcal{F}_i$s of an input sample into a bipolar value vector $\mathcal{V}_{f_i}$. Followed is the feature layer, which is essentially a 
sparse binary neural network to bind the bipolar feature values $\mathcal{V}_{f_i}$ with the corresponding feature vectors $\mathcal{F}_i$ through the Hadamard product. 
The last class layer equivalently performs similarity checking, where the weights of the layer are collections of all class vectors. This layer  outputs the score product for each class, and the class label with the highest score is taken as the classification result $\hat{y}$. It is worth noting that the inference in an LDC classifier is fully binary as the non-binary weights of ValueBox is not needed after training.
Compared to HDC, LDC classifiers have been demonstrated as a more promising alternative due to its lightweight model and high inference accuracy to support intelligent agents in the tiny devices.

\section{Problem Setup} 
We focus on the few-shot supervised learning in this work. Supervised learning learns a model that maps input data points $x \in \mathcal{X}$ which have a true label $y \in \mathcal{Y}$ to predictions $\tilde{y}$.  A task $\Tau_i$ is composed of $(\mathcal{X}, \mathcal{Y}, L, q)$, where $L$ is the task-specific loss function and $q$ is the data distribution of $\Tau_i$. We assume all data points are drawn i.i.d. from $q$.

Given the distributions over a set of tasks $p(\Tau)$, we aim at learning a general representation function $f(\cdot;\theta)$ using a handful of data points from each class. The $f(\cdot;\theta)$ is essentially a representation learning network parameterized by $\theta$, which can then fast adapt to previously unseen tasks in new devices by learning another adaption function $g(f(\theta);\phi)$ with little local data examples and (closed-form) gradient updates on  $\phi$ with minimal computation. 
In a nutshell, we propose to train the LDC backbone as a reusable template
to fast adapt to new tasks on tiny devices in the end. 

With the goal of obtaining a good initialization, we train the LDC model in a meta-learning manner. Our \ouralg consists of two phases: meta-training and meta-testing (also referred to as fast adaption). We meta-train on $m$ tasks $S_i\sim p(\Tau)$, $i=1,\cdots,m$ to learn the representation, and meta-test on a \textit{different} task $T\sim p(\Tau)$. 

In the training process of \ouralg, we introduce two updating loops as shown in our Algorithm~\ref{algo:meta-train}. In a nutshell, the inner loop updates model parameters with respect to an individual task using $K$ data points by one (or few) gradients steps, while the outer loop updates the entire model's parameters with regard to the loss after the inner loop updates.

In the fast adaption stage, we apply $M$-shot $N$-way evaluation,
where $N$ is the number of classes per task. We would use $M$ data examples from each class of the new task to update the partial model, and then carry out the evaluation on the testing dataset from the new task.

\section{The Design of \ouralg}
In this section, we present our architecture, referred to as the \ouralg. It uses a
carefully-crafted interleaved training algorithm 
to train the LDC classifier. The primary objective is to achieve fast adaption to unseen (but related) tasks on edge tiny devices by updating
partial learned model parameters with minimal computational cost. We provide an overview of the \ouralg in Figure~\ref{fig:lol}.

\subsection{Meta Training}

The original MAML~\cite{finn2017modelagnostic} takes the \textit{training-and-fine-tuning} pipeline, which optimizes $\theta$ and $\phi$ together in both meta-training and fast adaption. 
But, updating the non-binary weights $\theta$ can be too computationally expensive
for tiny devices.
Thus, in \ouralg, 
we instead use the \textbf{training-and-probing} pipeline. The
key idea of \ouralg is to separate the representation function $f(\cdot;\theta)$ and the prediction function $g(\cdot;\phi)$ for meta-training and fast adaption on tiny devices.  
Thus, we only optimize the representation function $f(\cdot;\theta)$ in the meta-training, and optimize the prediction function $g(\cdot;\phi)$ in fast adaption.

Specifically, in the meta-training, for each update step, we first learn a randomly initialized prediction function $g(\cdot;\phi)$ to classify examples based on a given representation $f(\cdot;\theta)$. An updated parameter $\phi_i$ is obtained using $K$ examples from the sampled tasks $S_i$ through one (or more) gradient steps w.r.t. the loss on the sampled tasks.  We then resample $K$ new examples from each class in $S_i$ and optimize the whole model w.r.t. $(\theta, \phi)$ across those tasks from $p(\Tau)$. The full algorithm is outlined in the Algorithm~\ref{algo:meta-train}, where $\alpha$, $\beta$ represent tunable step size. Intuitively, the $\phi$ serves as the task-specific embedding, which modulates the behaviour of the model. In the outer loop updates, by considering how the errors on specific task changes with respect to the updated model parameters, we expect to obtain a model initialization such that small changes in the model parameters could lead to substantial performance improvement for any task. This improvement has been empirically attested in our evaluations of section 5.

\begin{center}
\begin{algorithm}[ht]
\caption{\ouralg --- Training}
\label{algo:meta-train}
\begin{flushleft}
{\bfseries Input:} $\Tau$: the whole task set; $t$: number of outer gradient steps; $m$: Number of inner gradient steps (\textit{i.e.}, number of sampled meta-training tasks); $\alpha$, $\beta$: step size parameters.\\
{\bfseries Output:} $\theta$, $\phi$
\end{flushleft}
\begin{algorithmic}[1]
\State Randomly initialize parameters $\theta, \phi$
\For{$j$ in 1, 2, ..., $t$} {\color{blue}{\Comment{outer loop 
}}}
\For{$i$ in 1, 2, ..., $m$} {\color{orange}{\Comment{inner loop
}}}
\State Sample batches $B_i$ \Comment{each batch $B_i$ contains $K$ examples for each class in $S_i \sim p(\Tau)$}  
\State Derive task-specific $\phi_i'$: 
$\phi_i' \leftarrow \phi - \alpha \nabla_{\phi} L (B_i;\theta, \phi)$
\State Re-sample another batch $B'_i$ of the same batch size
\EndFor
\State Update both parameters $\theta$, $\phi$:
$(\theta, \phi) \leftarrow (\theta, \phi) - \beta \nabla_{\theta, \phi} \sum_{i=1}^{m} L (B'_i;\theta, \phi_i')$

\EndFor
\State Return $\theta$, $\phi$
\end{algorithmic}
\end{algorithm}
\end{center}

It is also worth noting that as the meta training involves potentially many tasks to learn a good initialization, the whole process of meta-training can be done on either the GPU or a powerful CPU. In contrast, adaptation is performed on local tiny devices with minimal computation cost as we explained in the subsequent subsections.

\subsection{Fast Adaption}
Given the learned initialization, we can fast adapt to a new task $T_i$ under a $M$-shot $N$-way setup. To adapt to a new task on each tiny device, we freeze the parameters of the representation network $\theta$ and only update the class layer's parameter $\phi$ by using only a few samples. By this means, we  preserve the broad knowledge learned from various tasks in the representation network $f(\cdot;\theta)$. Moreover, updating the last layer $\phi$ rather than the entire model is far less costly and therefore more affordable for tiny devices which are usually with stringent resource constraints.  Furthermore, we use the hinge loss instead of the commonly used cross-entropy for gradient updates, as the former requires less complicated arithmetic operations. Instead of requiring the model to compute gradients by itself, we feed the gradients of the hinge loss in a closed-form to the model directly, via Eqn. (1):

\begin{equation}
\small
  \nabla_{w_{j}} L_i =
    \begin{cases}
      - \sum_{j \neq y_i}  \mathbbm{1}  (w_j^T x_i - w_{y_i}^T x_i + \Delta > 0), &  j = y_i\\
      \mathbbm{1} (w_j^T x_i - w_{y_i}^T x_i + \Delta > 0)\cdot x_i, & j \neq y_i,
     
    \end{cases}       
\end{equation}
 where $\Delta$ is the desired margin; $w_{y_i}$ is the parameters corresponding to the correct class; $w_j$ is the rest of parameters, and $x_i$ is the vector representation for the data example $i$. We outline the fast adaption process in the Algorithm~\ref{algo:meta-test}. 
 
Interestingly, compared to fine-tuning entire parameters of all layers in the fast adaption, our experiments results have shown the accuracy achieved by the \ouralg, which only updates last layer's parameters, is not lower. This can be attributed to the alleviated meta over-fitting issue, as updating the entire model using a few data points from a task in the adaption stage can easily cause the model to overfit to this partial data distribution.

\begin{center}
\begin{algorithm}[t]
\caption{MetaLDC - Fast adaption to a new task $T$}
\label{algo:meta-test}
\begin{flushleft}
{\bfseries Input:} 
$\theta$, $\phi$: learned model parameters;
$t$: Number of gradient steps; $\gamma$: step size parameters.\\
{\bfseries Output:} $\phi$
\end{flushleft}
\begin{algorithmic}[1]
  \State Sample $M$ examples for each class of $T$
  \For{$j$ in 1, 2, ..., $t$}
  \State Update $\phi$ by: $\phi \leftarrow \phi - \gamma \nabla_\phi L(\cdot;\theta, \phi)$
  \EndFor
\State Return $\phi$
\end{algorithmic}
\end{algorithm}
\end{center}
\setlength{\textfloatsep}{0pt}

\section{Evaluation}
In this section, we empirically evaluate the performance of \ouralg compared to several baselines on both vision and non-vision datasets. We have considered various performance perspectives, including adaption accuracy on unseen tasks with a small amount of data points, robustness against hardware random bit errors, the latency and hardware energy consumption, as well as the change of model's accuracy w.r.t. the hyperparameters values selection. Additionally, we have also evaluated the efficacy of the learned representation $f(\cdot;\theta)$.

\begin{figure*}[t]
\centering
\subfigure{\label{fig:long_seq}
\tabcolsep 2pt
\adjustbox{valign=c}{
    \resizebox{0.55\textwidth}{!}{
    \begin{tabular}{lcccccc}
    \toprule
    & & \multicolumn{5}{c}{Evaluation Tasks} \\
    \cmidrule(lr){3-7}
    Methods & $K$-shot & \multicolumn{1}{c}{$T_1: 0^{\circ}$} &   \multicolumn{1}{c}{$T_2: 2^{\circ}$} &  \multicolumn{1}{c}{$T_3: 4^{\circ}$} &  \multicolumn{1}{c}{$T_4: 6^{\circ}$} &  \multicolumn{1}{c}{$T_5: 8^{\circ}$} \\
    \midrule
    HDC w/ retraining\quad & - & \acc{76.74}{0.74} & \acc{77.75}{0.65} & \acc{77.01}{0.57} & \acc{77.87}{0.44} & \acc{79.13}{0.53} \\
    Pretrained LDC & - & \multicolumn{1}{c}{\acc{78.25}{2.37}} & \multicolumn{1}{c}{\acc{78.26}{3.03}} & \multicolumn{1}{c}{\acc{78.67}{2.97}} & \multicolumn{1}{c}{\acc{79.25}{2.50}} & \multicolumn{1}{c}{\acc{80.89}{1.76}}  \\
    \midrule
    \multirow{2}{*}{\ouralgfull\quad} & $K=1$ & \acc{80.03}{0.84}  & \acc{81.01}{0.57} & \acc{82.37}{0.81}  & \acc{84.04}{0.47} & \acc{85.34}{0.23} \\
    & $K=5$ & \acc{80.35}{0.66} & \acc{81.78}{0.53} & \acc{83.25}{0.57} & \acc{84.97}{0.18} & \acc{86.31}{0.05} \\
    \cmidrule{2-7}
    \multirow{2}{*}{\textbf{\ouralg}}\quad & $K=1$ & \color{blue} \textbf{\acc{82.78}{0.97}} & \color{blue}  \textbf{\acc{82.99}{1.07}}   &
    \color{blue}\textbf{\acc{84.84}{1.22}} &
    \color{blue}\textbf{\acc{86.33}{0.55}} & \color{blue}\textbf{\acc{87.97}{0.42}}  \\
    & $K=5$ &  \color{red} \textbf{\acc{82.83}{0.71}} & \color{red} \textbf{\acc{83.11}{0.58}} &
    \color{red} \textbf{\acc{85.74}{0.83}} &
    \color{red} \textbf{\acc{86.54}{0.23}} & \color{red} \textbf{\acc{88.01}{0.36}} \\
    \midrule
    \multirow{2}{*}{Upper Bound} & $K=1$ &  \acc{87.00}{0.54} &  \acc{87.69}{0.42} &  \acc{88.77}{0.33} &  \acc{90.19}{0.33} &  \acc{92.94}{0.19} \\
    & $K=5$ & \acc{87.53}{0.43} &\acc{87.74}{0.39} & \acc{89.01}{0.27} & \acc{90.87}{0.29} & \acc{93.11}{0.11} \\
    \bottomrule
    \end{tabular}
    }
    }
   
}
\subfigure{\label{fig:analyse_b}\includegraphics[width=0.40\textwidth, valign=c]{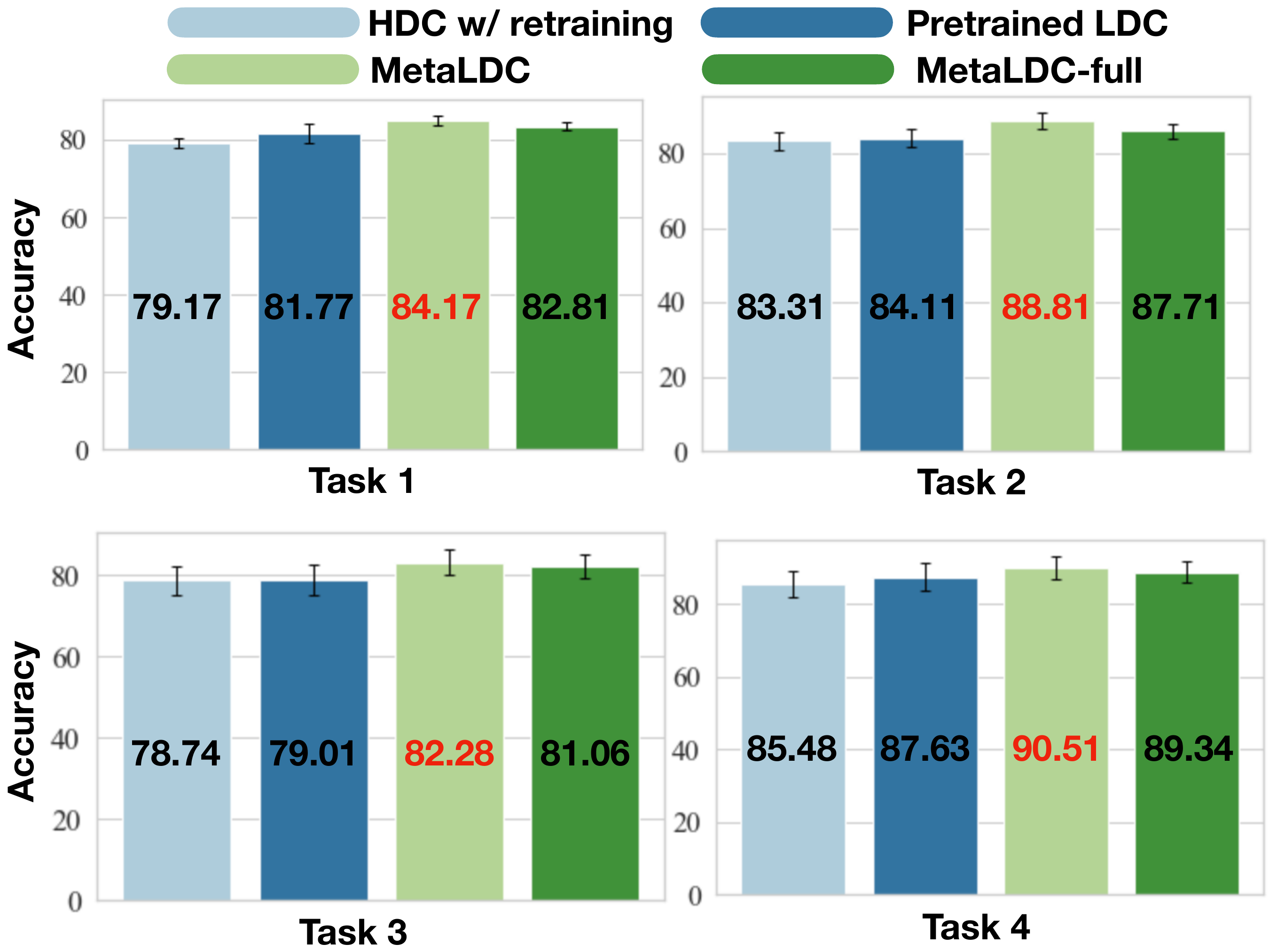}}
\caption{Left: Accuracy of MetaLDC compared to other methods on the Rotated MNIST. Right: Accuracy of MetaLDC compared to other approaches
on the Split-ISOLET. {\color{red}{Red}} marks the results of the highest accuracy, and {\color{blue}{blue}} marks the results of the second highest.}
\label{main-res-fig}
\end{figure*}

\subsection{Setup}
We describe the datasets and tasks, baselines, as well as implementation details here.
\paragraph{Datasets}
We use two benchmarking lightweight datasets, for both vision and non-vision applications, to evaluate \ouralg  on tiny devices. One is the Rotated MNIST, which is originated from MNIST~\cite{MNIST}. In Rotated MNIST, each image contains the digit rotated by a certain degree. The other is referred to as Split ISOLET, derived from the UCI ISOLET dataset~\cite{uci}. The ISOLET contains 26 classes in total, and we divide them into separate tasks, where each task contains 4 unique randomly sampled classes. 

\paragraph{Training and Evaluation Tasks} In the Rotated MNIST, the rotation degree of tasks we use to train the methods are between $[10^{\circ}, 20^{\circ})$. 
The learned models are then evaluated on the testing dataset of Rotated MNIST with rotation degree from $\{0^{\circ}, 2^{\circ}, 4^{\circ}, 6^{\circ}, 8^{\circ}\}$. For the Split ISOLET, we use 20 classes from the ISOLET to generate the training tasks, while the remaining classes form the candidate classes pool to produce evaluation tasks.

\paragraph{Baselines} We compare \ouralg with both HDC methods and other LDC-based models. In the \textit{Pretrained LDC} method, we pretrain the LDC model with standard supervised learning using the whole training datasets from the training tasks. Then we also use the Algorithm~\ref{algo:meta-test} for fine-tuning. Another baseline we designed is the \textit{\ouralgfull}. The training algorithm of \ouralgfull is the same as \ouralg, as shown in the Algorithm~\ref{algo:meta-train}. The difference is in the fine-tuning stage, where we update the parameters of the entire model, not only the last layer for the \ouralgfull. Note that \ouralgfull is prohibitively expensive for tiny devices, as it requires keeping the entire weights of the LDC model and full backpropagation calculations throughout the fast adaptation process.
Besides the LDC variants, we also compare with a SOTA HDC method, the \textit{HDC with retraining}. The HDC w/ retraining would give more weights to a misclassified sample in its correct class hypervector and subtracted from the wrong class hypervector in training to improve the HDC classification accuracy~\cite{Imani2020QuantHDAQ}. We set $D=8,000$ for the HDC models in our experiments following the setup of~\cite{ldc}. In addition, 
we put the multi-layer perceptrons (MLP) here as an upper bound, although it is not feasible to be deployed on extremely resource-constrained tiny devices. In the MLP, we use Algorithm~\ref{algo:meta-train} for training and Algorithm~\ref{algo:meta-test} for fast adaption.

\begin{figure}[ht]
    \centering
\includegraphics[width=0.8\columnwidth]{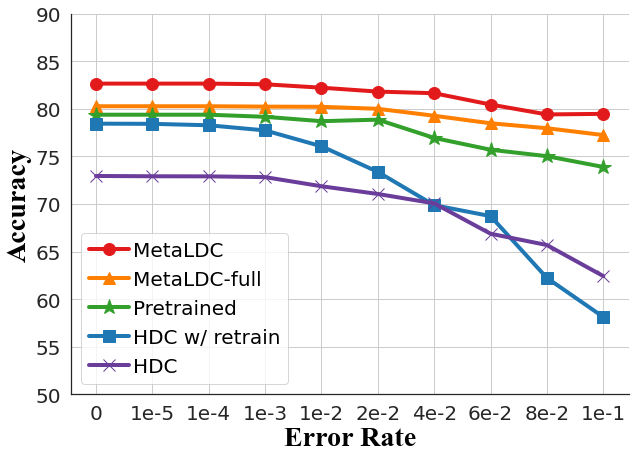}
    \caption{Bit error robustness of different models on the MNIST testing dataset.}
    \label{fig:robust}
\end{figure}

\paragraph{Implementation Details}
In the meta training stage, for the Rotated MNIST, we train the model for $60$ epochs with a  batch size of $10$. In the Split ISOLET, we use $30$ epochs w.r.t. its smaller dataset size. In the Rotated MNIST experiments, we used $K=\{1, 5\}$ for gradient updates in the meta learning involved methods, while setting $K=1$ in the Split ISOLET. We train all models with the Adam optimizer~\cite{adam} except the HDC classifier, which is not feasible to fit in any standard training  optimizer.

In the fast adaption stage, we set $M = 10$ for the Rotated MNIST and $M = 5$ in Split ISOLET, to update $\phi$. To ensure fair comparison, the same data points are used across different methods. Note that we don't fine-tune the HDC classifier as there seems no feasible way to update the learned hypervectors which are essentially composed of zeros and ones for the new incoming data examples.

\subsection{Testing Accuracy}
We show our evaluation results on the Rotated MNIST in Figure~\ref{main-res-fig}. From Figure~\ref{main-res-fig}, we can see that \ouralg outperforms other baselines across all evaluation tasks. We observe that \ouralg has achieved higher accuracy compared to the \ouralgfull, which updates the entire model parameters to adapt to a new task in the fine-tuning stage. We attribute this to over-fitting, as the entire model focuses on learning a very small amount of data from the task. In comparison, \ouralg, which only updates the last layer while keeping the former layers untouched, has alleviated this over-fitting issue to certain extent. We can also observe that as the rotation degree of the evaluation data becomes larger, the testing accuracy of the \ouralg also increases due to higher similarity between the training and the evaluation tasks.

For the Split ISOLET, the evaluation results are reported in the Figure~\ref{main-res-fig}. Based on Figure~\ref{main-res-fig}, the accuracy achieved by \ouralg is the highest on different tasks. The second highest is the \ouralgfull, which is not as computationally efficient as the \ouralg for tiny devices.

\subsection{Robustness against Hardware Bit Errors}
One of the most appreciated merits of the HDC-based models is the robustness against random bit errors on the hardware. The plain LDC model has been shown that it can achieve comparable robustness by the uniformly distributed information in the compact vector representation, given the largely reduced dimensionality~\cite{ldc}. Our empirical results in Figure~\ref{fig:robust} have shown that \ouralg exhibits even stronger robustness than the pretrained-then-finetuned LDC model.
This can be attributed to a more general representation learned via our meta training algorithm. The broad prior knowledge stored in the reusable template helps in adapting to tasks with noise perturbations and fighting against the random bit errors. Compared to the other methods, \ouralg has also shown slower decline of accuracy.

\subsection{Inference Cost on Hardware}
Here, we evaluate the inference efficiency of \ouralg and HDC w/ retraining following the same hardware pipeline setup as in the~\cite{ldc}. The hardware platform we use is the Zynq UltraScale+, where we transform the bipolar values \{1, -1\} to \{0, 1\} in the implementation. In the experiment, we limit the resource usage (e.g., lookup table (LUT) < 10k) to approach the common practice in tiny devices. 

We report the numerical results in the Table~\ref{tab:cost}. From the Table~\ref{tab:cost}, we can see that the LDC models are at least $100x$ faster than HDC w/ retraining classifiers. The model size based on LDC is $150x$ smaller than the HDC ones. The energy consumption of the MAML LDC (tiny) are less than $100nJ$ in the evaluation datasets, which has demonstrated great improvement on hardware acceleration compared to the HDC-based models. Note we don't measure the MLP-based model cost here, as its inference requires matrix multiplication via floating-point operators rather than simple binary arithmetic, making it too resource-intensive to run on a tiny device. On top of that, the MLP architecture cannot be trivially supported by the FPGA platform due to the floating point computation~\cite{cost3}, and inter-platform comparison of algorithm performance is considered neither instructive or fair. Even though there is MLP with the fix-point format which could be implemented on FPGA, the required utilization of DSP and other resources are still tremendous to carry out the involved matrix multiplication, far above the resource budget of a tiny device.

\begin{table}[h]
  \small
  \caption{Inference cost comparison between \ouralg and the HDC w/ retraining on the Zynq UltraScale+.
  }
    \centering
    \resizebox{\columnwidth}{!}{%
    \begin{tabular}{cc ccc}
    \toprule
    \textbf{DataSet} & \textbf{Model}  & \textbf{Size (KB)} & \textbf{Latency (us)} & \textbf{Energy (nJ)}       \\
    \midrule
     \multirow{2}{*}{R-MNIST}& \textbf{\ouralg} & \textbf{6.48} & \textbf{3.99} & \textbf{64}\\
    & HDC  & 1050 & 499 & 36926\\  
    \midrule
     \multirow{2}{*}{S-ISOLET} & \textbf{\ouralg} & \textbf{5.10} & \textbf{3.13} & \textbf{38}\\
    & HDC  & 877 & 388 & 29488\\ 
    \bottomrule
    \end{tabular}
    }
    
    \label{tab:cost}
\end{table}


\begin{figure*}[ht]

\centering
\subfigure[]{\label{fig:a}\includegraphics[width=0.245\textwidth]{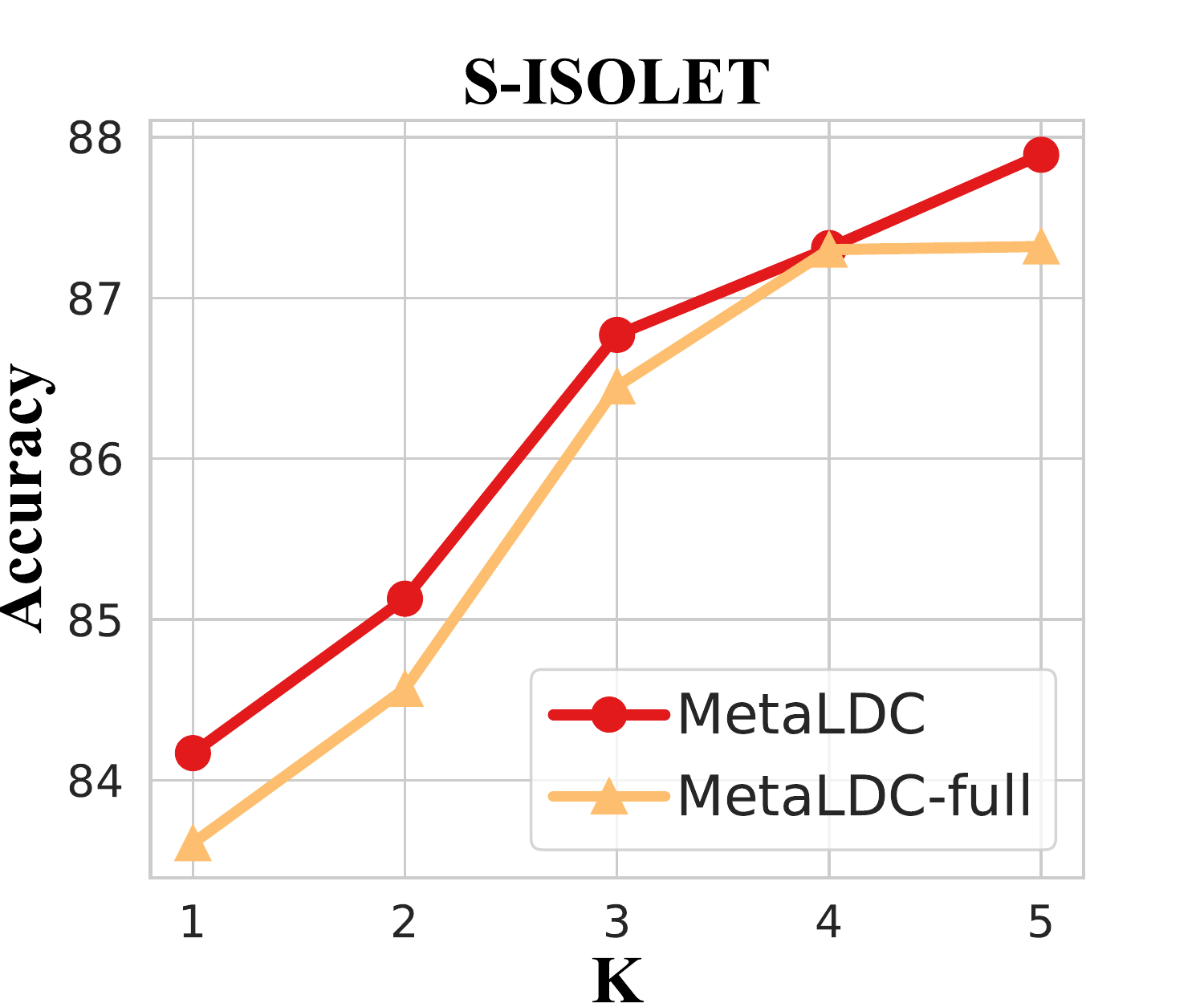}}
\subfigure[]{\label{fig:b}\includegraphics[width=0.245\textwidth]{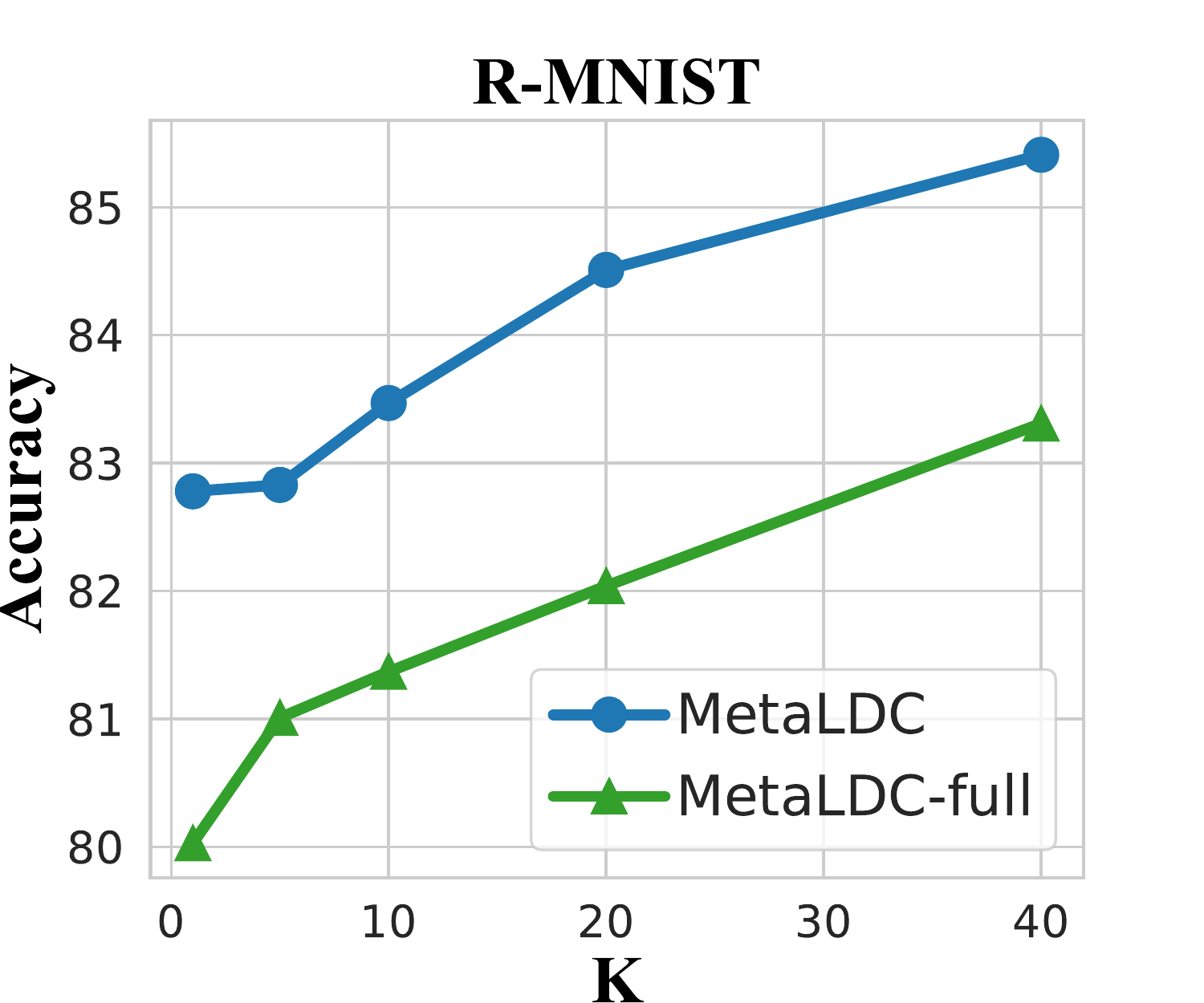}}
\subfigure[]{\label{fig:c}\includegraphics[width=0.245\textwidth]{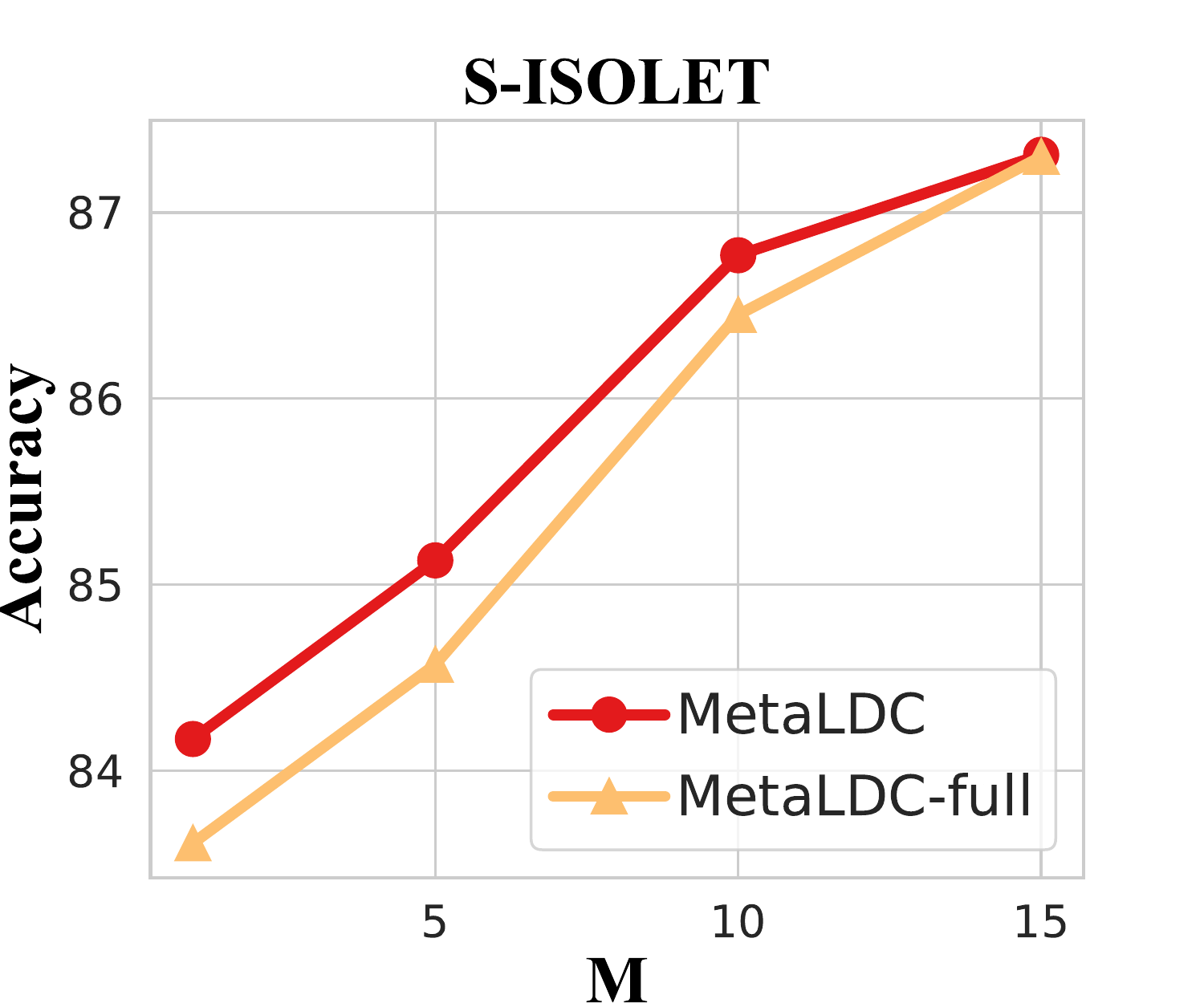}}
\subfigure[]{\label{fig:d}\includegraphics[width=0.245\textwidth]{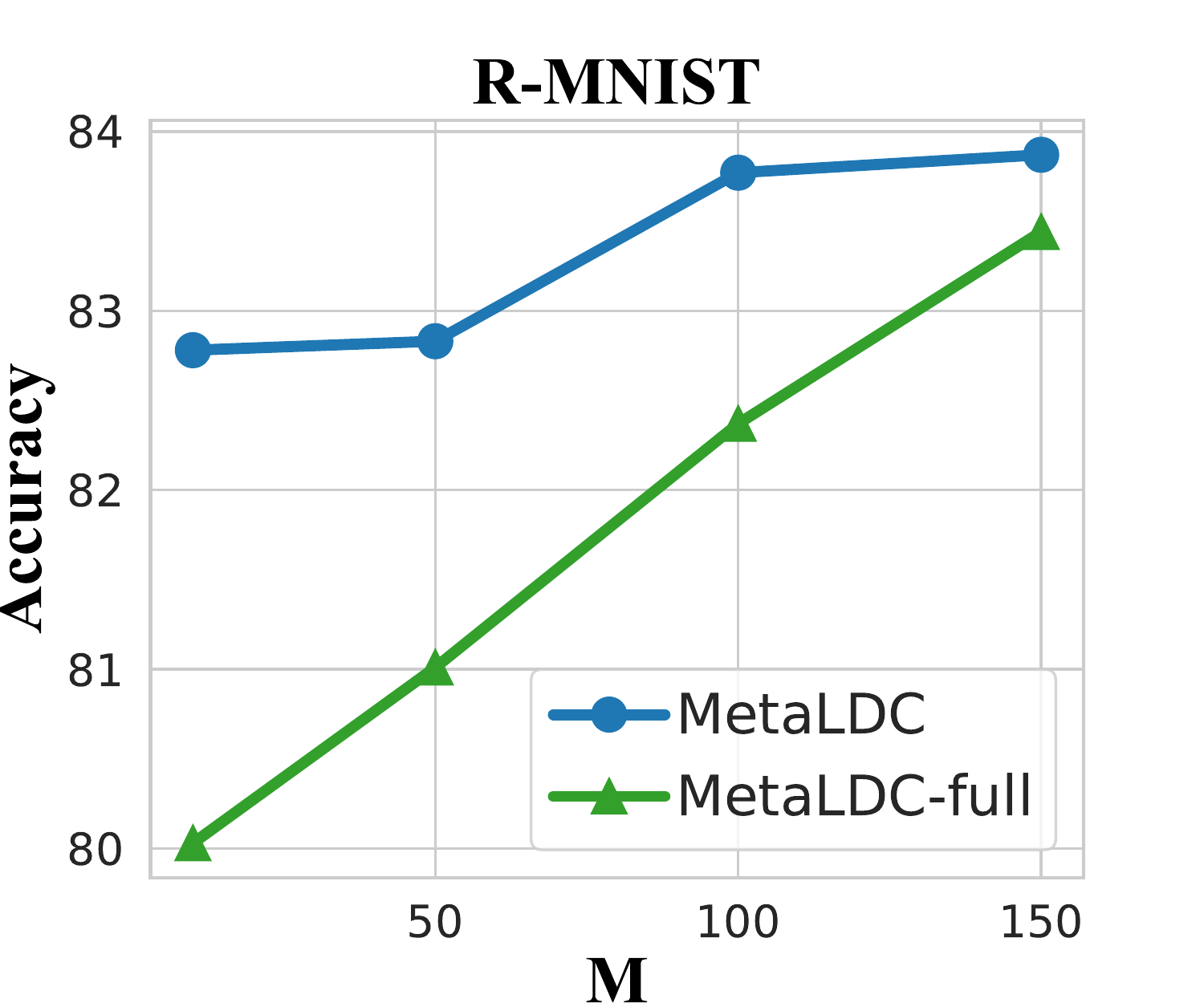}}
\caption{Hyper-paramters ablation study on different datasets w.r.t. K, the number
of examples sampled from each class in the meta-training;
and M, the number of examples sampled from each class of a new task
in the fast adaption.}
\label{fig:abl}
\end{figure*}

\subsection{Hyper-parameter Ablation Studies}
To showcase the performance of \ouralg with different choice of hyperparameters, we conduct the ablation study for $K$, the number of data examples to train the model; and $M$, the number of samples we used from each class of a new task to update $\phi$ in the fast adaption stage. Depending on different dataset size, for the Rotated MNIST, we set the $K=\{1, 5, 10, 20, 40\}$, while $K=\{1, 2, 3, 4, 5\}$ for the Split-ISOLET dataset. The $M$ is set as $\{10, 50, 100, 150\}$ in the Rotated MNIST, and $\{1, 5, 10, 15\}$ in the Split ISOLET, respectively. The evaluation dataset we used in the Rotation MNIST is the the original MNIST, while the images in the training data have rotation degree in $[10, 20)$. In the Split-ISOLET, we use the Task 1 as the evaluation task. 

The results are provided in Figure~\ref{fig:abl}. We observe that as the values of $K$ and $M$ increase, the accuracy of both \ouralg and \ouralgfull has improved. We also notice that the accuracy gap between \ouralgfull and \ouralg become smaller as $K$ or $M$ increase, especially on the Split ISOLET dataset. We attribute the performance increases of \ouralgfull to larger percentage of data points sampled from the task to update the model, which gradually diminishes the over-fitting effect.

\subsection{Effectiveness of the Learned Representation}
\begin{figure}
    \centering
\includegraphics[width=\columnwidth]{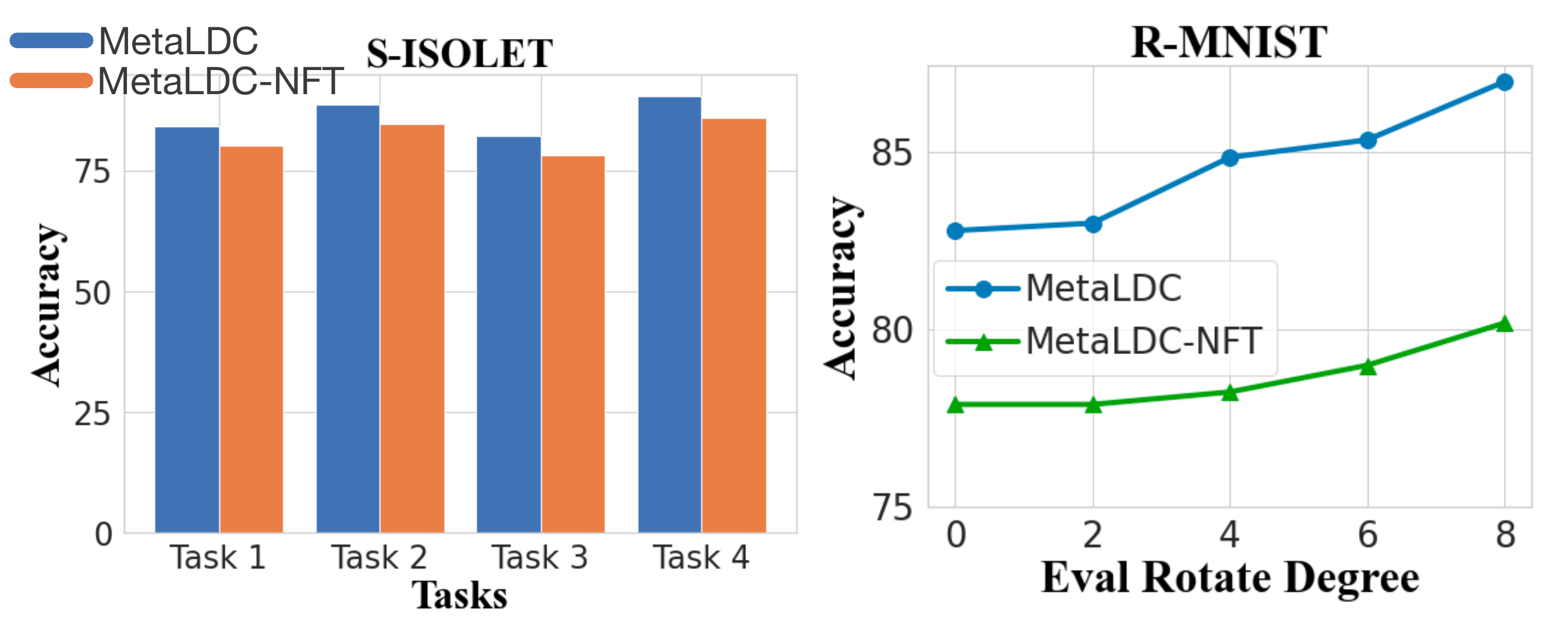}
    \caption{Accuracy comparison between \ouralg and \ouralgnft on S-ISOLET and R-MNIST with $K=1$.}
    \label{fig:suppbase}
\end{figure}

To study the efficacy of learned representation by our meta-training process, we have designed another method, referred to as the \ouralg-NonFineTuning (\ouralgnft). In this method, we use the Algorithm 1 to train the LDC model. We then test its accuracy on the new task \textit{without} any fine-tuning. As shown in the Figure~\ref{fig:suppbase}, we can see that the accuracy gap between \ouralgnft and \ouralg is within $5\%$ on the Split ISOLET and $10\%$ on the Rotated MNIST, which has reflected our Algorithm 1 has produced a good initialization to some extent.

\section{Related Works}
\textbf{Meta learning.} By distilling the learning experience from a broad set of related tasks, MAML~\cite{finn2017modelagnostic} has achieved great success in the fast adaption regime~\cite{citemaml1, citemaml2, citemaml3, citemaml4, citemaml5}.
The \cite{lin2020collaborative, fedmeta2} have proposed distributed collaborative frameworks to leverage knowledge between edge nodes via MAML. 
To reduce the computational cost, the~\cite{meta-edge} has presented a divide-and-conquer approach where the linear approximation is utilized to estimate the Hessian, while \cite{meta-embedded} has discussed using MAML with synthetic gradients in a feed forward manner for deep neural networks. However, most approaches  are still quite costly, not viable for tiny devices with severe resource constraints. 

\textbf{Hyper-dimensional computing.} HDC has been known as an efficient alternative to expensive deep neural networks for tiny devices \cite{HDC_Survey_Review_IEEE_Circuit_Magazine_2020_9107175,HDC_VSA_Encoding_ZhiruZhang_Cornell_arXiv_2022,Najafabadi2016HyperdimensionalCF,duan2022lehdc,quant-hdc1,quant-hdc2, basaklar2021hypervector, hhdc}. The study \cite{quant-hdc2} has proposed to use vector quantization to further reduce the model size. Besides, some works have optimized HDC's encoding and training to improve its accuracy on a single data distribution~\cite{adaptHDC}. Nonetheless, the required HDC model size to
obtain an acceptable accuracy is still prohibitive large for tiny
devices.
More recently, LDC has been studied to significantly
improve the efficiency of HDC \cite{ldc}, where the encoded vectors
are only tens, yet the accuracy is even
higher.
Nevertheless, fast adaption problems to unseen but related tasks have not been well addressed
in either HDC or LDC. 

\textbf{On-device learning.} The fast adaption issue has become even more pressing
for edge tiny devices due to their low latency tolerance
and limited computational power~\cite{tiny1, tiny2, tiny3, tiny4, Luo_2017_ICCV}. For example, FSCL~\cite{FSCL} and subsequent C-FSCIL~\cite{C-FSCIL} also utilize the frozen meta-learned network structure for adaption. However, they focus on addressing the catastrophic forgetting issue by storing the hyperdimensional prototypes of past classes in the online class incremental setting, rather than efficient edge device computing or adaption. Meanwhile, their replay buffer composed of hyper-dimensional vectors is costly if deployed to tiny devices.
In~\cite{tiny}, the MCUNet is proposed to find the optimal neural architecture by neural architecture search with resource constraints of heterogeneous tiny devices, whereas our \ouralg does not require additional search efforts but provides a reusable lightweight template for unseen tasks.

\section{Conclusion}
In this paper, we propose \ouralg, a LDC-based approach to fast adapt to unseen tasks via interleaved meta training  for resource-constrained tiny devices. In \ouralg, the LDC architecture is first trained across a set of different tasks, where we separately train the task-specific parameters $\phi$ in the inner loop of the meta-training algorithm. The learned model can then fast adapt to a new task by only updating the last layer using a handful of data points, while preserving learned prior knowledge in the former layers. Our empirical results have shown that our method has achieved higher accuracy compared to the HDC methods and pretrained LDC classifiers. 

\textbf{Limitations \& Future Work} Our work has made the first step towards leveraging a novel low-dimensional classifier in on-device transfer learning for tiny gadgets and appliances. Further investigation on the feasibility of using other low complexity non-LDC NNs on tiny devices, as well as a more comprehensive comparison between them and the \ouralg are desired in future works. Besides, extending \ouralg to the unsupervised learning setting is a natural future direction to explore.

{
\bibliographystyle{plain}
\bibliography{references}
}

\end{document}